\documentclass[11pt]{article}

\usepackage[final]{acl}

\usepackage{times}
\usepackage{latexsym}

\usepackage[T1]{fontenc}

\usepackage[utf8]{inputenc}

\usepackage{microtype}
\usepackage{inconsolata}
\usepackage{fontawesome}
\usepackage{hyperref}
\usepackage{booktabs}
\usepackage{svg}

\usepackage{listings}

\usepackage{graphicx}

\makeatletter
\def\@fnsymbol#1{\ifcase#1\or *\or $\dagger$\or $\ddagger$\or 
   \mathsection\or \mathparagraph\or \|\or **\or $\dagger$\dagger$
   \or $\ddagger$$\ddagger$ \else\@ctrerr\fi}
\makeatother

\title{Dr.Copilot: A Multi-Agent Prompt Optimized Assistant for Improving Patient-Doctor Communication in Romanian}

\author{
 \textbf{Andrei Niculae\textsuperscript{1,$\dagger$}},
 \textbf{Adrian Cosma\textsuperscript{1,2,$\dagger$}},
 \textbf{Cosmin Dumitrache\textsuperscript{3,$\ddagger$}},
 \textbf{Emilian Rǎdoi\textsuperscript{1,3,$\ddagger$}},
\\
 \textsuperscript{1}National University of Science and Technology POLITEHNICA Bucharest,\\
 \textsuperscript{2}Dalle Molle Institute for Artificial Intelligence Research (IDSIA),\\
 \textsuperscript{3} MedicChat
\\
 \small{
   \textbf{Correspondence:} \href{mailto:emilian.radoi@upb.ro}{emilian.radoi@upb.ro}
 }
}

\begin{document}

\maketitle

\begin{abstract}
Text-based telemedicine has become increasingly common, yet the quality of medical advice in doctor-patient interactions is often judged more on how advice is communicated rather than its clinical accuracy. To address this, we introduce Dr.Copilot , a multi-agent large language model (LLM) system that supports Romanian-speaking doctors by evaluating and enhancing the presentation quality of their written responses. Rather than assessing medical correctness, Dr.Copilot provides feedback along 17 interpretable axes. The system comprises of three LLM agents with prompts automatically optimized via DSPy. Designed with low-resource Romanian data and deployed using open-weight models, it delivers real-time specific feedback to doctors within a telemedicine platform. Empirical evaluations and live deployment with 41 doctors show measurable improvements in user reviews and response quality, marking one of the first real-world deployments of LLMs in Romanian medical settings.
\end{abstract}

{\renewcommand{\thefootnote}{\fnsymbol{footnote}}{
\footnotetext[2]{Equal contribution.}
\footnotetext[3]{Equal supervision.}
}}

\section{Introduction}
\begin{figure}[t!]
    \centering
    \includesvg[width=1.0\linewidth]{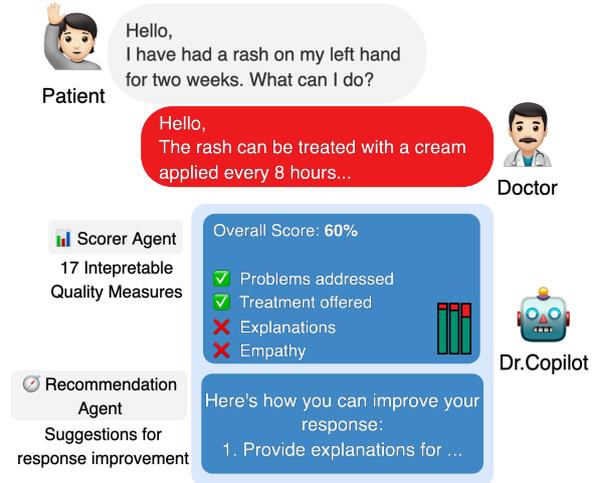}
    \caption{Dr.Copilot evaluates a doctor’s response to a patient's question \textit{in Romanian}\protect\footnotemark, identifying areas for improvement and offering suggestions to improve the form of the response.}
    \label{fig:teaser}
\end{figure}

Since 2020, telemedicine services have experienced rapid growth and widespread adoption \cite{peters2024telemedicine}. With social distancing measures and lockdowns limiting access to in-person healthcare, virtual consultations quickly became a practical and convenient alternative, offering easier and faster access to doctors. Doctors providing health advice, especially in the case of text-based telemedicine services, have to balance two competing priorities: on one hand, they have to provide the most medically accurate advice possible, and on the other, they have to communicate this advice in a way that is most helpful to patients. Generally, doctors focus on the first but sometimes tend to neglect the second. Unfortunately, this tends to attract negative reviews as the patient's perception on quality is formed primarily by how well the information is presented \cite{martin2005challenge,reis2024influence}. This is simply due to most patients not being able to judge the quality from a medical perspective.

\footnotetext{For illustration, this example is translated into English.}

Although most telemedicine platforms that offer text-based services include an onboarding process with guidelines for doctors on how to present their medical advice to receive better user reviews, these recommendations are often overlooked, largely because they are delivered all at once, they tend to be too general, and are easily forgotten or ignored.

To help doctors better communicate medical advice on telemedicine platforms, we propose Dr.Copilot, a multi-agent assistant, designed to provide specific, case-by-case advice on how to improve the presentation / packaging of the medical responses in a way optimized for user satisfaction, without interfering with the medical content.
A well presented medical advice is more likely to be followed by the patient and have a positive impact \cite{martin2005challenge,reis2024influence}. From the doctors' perspective, a well presented response attracts positive reviews, directly impacting reputation and professional satisfaction \cite{street2009does,sargeant2008understanding}. From the telemedicine platform's perspective, better user satisfaction translates to retention and referral, which ultimately translates to better business outcomes such as sustainable growth and profits.

In this paper, we work on real-world data from a Romanian telemedicine platform that was used to optimize Dr.Copilot for this use case. Since the data is in Romanian, a low resource language \cite{nigatu-etal-2024-zenos}, it induces additional practical challenges.

Romanian is underrepresented in medical NLP research, and even more so in industry applications. In general, data is scarce and is even more lacking for specialized domains. Current Romanian LLMs \cite{masala2024rollama} and language resources \cite{masala-etal-2020-robert} can be useful for their general purpose embeddings, but might not be mature enough for direct use in real-world applications. However, the performance of multilingual LLMs \cite{huang2024survey,team2025gemma} is improving. Models such as Gemma \cite{team2025gemma} or its medical specialized counterpart MedGemma \cite{medgemma-hf}, have shown good multilingual performance and have robust results across many axes. Still, Romanian is not comprehensively evaluated, which induces inherent risks with direct human interaction.

Dr.Copilot is a multi-agent system designed to support Romanian-speaking doctors by evaluating their written responses to patient questions and offering targeted, constructive feedback (Figure \ref{fig:teaser}) across 17 interpretable measures of response quality \cite{mehri-eskenazi-2020-unsupervised,luis2024dialogue,shin2020generating}. We used three different LLM agents: a \textit{Scorer Agent}, a \textit{Recommender Agent} and, for evaluation, a \textit{Reconciliation Agent}. To minimize annotation efforts and unnecessary computational costs, we optimized the agents using an automatic prompt optimization approach, using DSPy \cite{khattab2024dspy}. We evaluate Dr.Copilot both in offline settings and in a live production environment with 41 doctors, one of the only real-world deployments of Romanian LLMs in telemedicine today.

We make the following contributions:
\begin{enumerate}
    \itemsep0em 
    \item We introduce Dr.Copilot, a multi-agent LLM system designed for Romanian text-based telemedicine that evaluates and enhances doctor-patient communication quality across 17 interpretable dimensions. We make the code publicly available\footnote{\href{https://github.com/nan-dre/dr-copilot}{\faicon{github} nan-dre/dr-copilot}}.
    \item We develop and validate an automatic prompt optimization approach using DSPy for medical communication assessment, achieving effective performance with limited labeled data (100 annotated examples) while ensuring privacy-preserving deployment through open-weight models.
    \item We show one of the first real-world deployments of LLM agents in Romanian medical setting, with measurable improvements in response quality and patient satisfaction (70.22\% increase in positive reviews) through live evaluation with 41 doctors.
\end{enumerate}

\section{Related Work}
\begin{figure*}[hbt!]
    \centering
    \includesvg[width=1.0\linewidth]{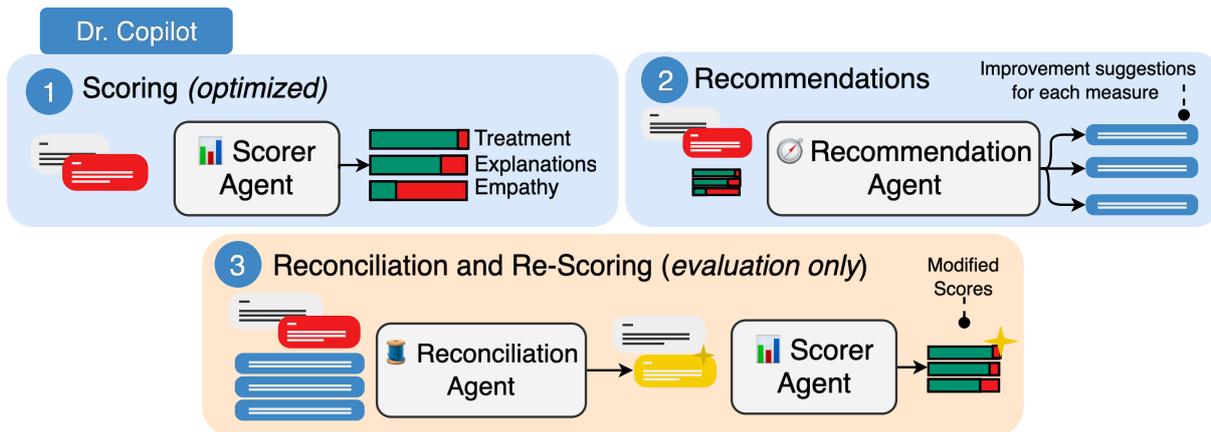}
    \caption{Overall diagram of Dr.Copilot. A patient-doctor interaction is scored and recommendations are generated for each measure. For automatic evaluation of recommendations we use a separate agent.}
    \label{fig:evaluation-architecture}
\end{figure*}
LLMs have gained traction in medicine \cite{zhou2023survey}, through both general-purpose models \cite{openai2024gpt4technicalreport,geminiteam2025geminifamilyhighlycapable} and domain-specific ones such as MedPalm \cite{singhal2025toward,singhal2023large}, MedGemma \cite{medgemma-hf}, and MedAlpaca \cite{han2023medalpaca}. In Romanian, smaller models such as RoLLama2/3, RoMistral, and RoGemma exist \cite{masala2024rollama}, but lack medical evaluations. Thus, large multilingual models remain viable alternatives \cite{team2025gemma,huang2024survey,isbister2021should}. LLM use in healthcare varies by model autonomy. Fully autonomous systems such as RiskAgent \cite{liu2025riskagent}, Polaris \cite{mukherjee2024polaris}, and Healthcare Copilot \cite{ren2024healthcare} manage full patient interactions \cite{zhao2025smart}, but raise legal and ethical concerns\footnote{\href{https://www.egattorneys.com/unauthorized-practice-of-medicine-bpc-2052}{\url{https://www.egattorneys.com/unauthorized-practice-of-medicine-bpc-2052}}, Accessed 04.07.2025}. Patients may receive biased responses \cite{kusa2023dr} due to model sycophancy \cite{sharma2023towards}, a side effect of RLHF \cite{rlhf}, reducing trust \cite{reis2024influence}. Safer alternatives restrict autonomy and advice generation \cite{cheng2024don}, supported by benchmarks such as HealthBench \cite{arora2025healthbench} and HeaL \cite{cheng2024don}.

As opposed to other works that design their application to directly offer medical advice in the absence of a medical professional \cite{li2023chatdoctor,zhao2025smart,zhao2025medrag}, we design our multi-agent system, Dr.Copilot, such that doctors can improve their communication with the patient, while having full control of the medical content. Specific to our use case, we focus on Romanian telemedicine, a novel application of a multi-agent LLM system to date.

\section{Method}
In designing Dr.Copilot, we were guided by several \textit{desiderata}: \textit{(i)} the feedback given to the doctor response should be structured across several easy to understand and interpretable axes;  \textit{(ii)} the feedback should be related only to presentation and not the medical content of the response; \textit{(iii)} there should be targeted, explicit and actionable improvement suggestions; \textit{(iv)} the doctor decides whether to incorporate the feedback or ignore the suggestions. 

One of the reasons for the reduced autonomy of Dr.Copilot is to avoid language misunderstanding due to the use of Romanian, which is not the primary language of current open-weight LLMs \cite{team2025gemma}. Moreover, patients lose trust in the medical advice if they suspect involvement of AI \cite{reis2024influence}, so doctors are supposed to formulate the content of the response in its entirety. We further detail the dataset construction for response scoring, experimental setup for prompt optimization and evaluation.

\subsection{Measuring the Quality of Doctor Responses}

To ensure control of responses from the model, we developed several measures of response quality, inspired by both dialogue evaluation literature \cite{mehri-eskenazi-2020-unsupervised,luis2024dialogue,shin2020generating} and several business-oriented quality measures. In Appendix \ref{sec:appendix}, we provide a description of each quality measure, including an empathy score \cite{shin2020generating}, grammatical score, scores related to specificity and relevancy of the response to the patient questions (problems addressed) \cite{gopalakrishnan2019topical,adiwardana2020towards}, and measures of how informative the response is (e.g., explanations for patient symptoms, causes, treatment) \cite{zhang2019dialogpt,guo2018topic}. Furthermore, we included measures directly related to the functionality of the telemedicine platform: inappropriate use of the clarifications functionality, or responding outside of the doctor's specialty. 

From the over 100,000 existing text consultations on the telemedicine platform, we randomly selected 100 question-response pairs that included both positive and negative user reviews. These pairs were annotated by two company employees, a male and a female, with business domain knowledge. Both individuals work in operations and support for the platform and, due to the nature of their roles, have explicit authorization to access patient data under strict confidentiality agreements. Given this context, no additional risk disclaimers for the annotators were necessary. All patients and doctors have agreed to have their questions and responses processed.
Since prompt optimization frameworks such as DSPy \cite{khattab2024dspy} require only a few ($< 100$) representative examples, the quality of annotations is more important than the quantity. For \textit{,,Empathy"} and \textit{,,Problems addressed"}, responses are scored on a 1-to-5 Likert scale \cite{likert1932technique}, while the rest are given a binary score. In the initial round of annotations, due to part of the metrics being subjective, the annotators had poor agreement (in terms of Cohen's Kappa \cite{cohen1960coefficient}), as shown in Table \ref{tab:agreement} (App. \ref{sec:appendix}). In a second round of annotations, the metrics were reviewed and clarified to ensure full agreement. Figures \ref{fig:metric-proportion} and \ref{fig:int-metric-proportion} (App. \ref{sec:appendix}) show the distributions for each measure. Since this dataset is proprietary and contains sensitive patient data, it will not be made public.

\begin{figure}[hbt!]
    \centering
    \includesvg[width=1.0\linewidth]{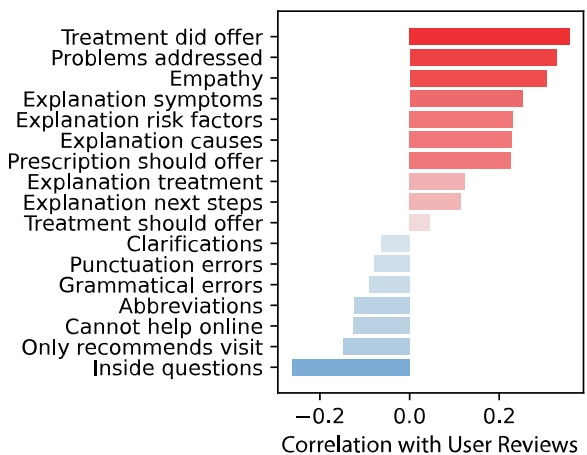}
    \caption{Correlation between each quality measure for doctor responses and user reviews.}
    \label{fig:metric-feedback-corr}
\end{figure}

In Figure \ref{fig:metric-corr} (App. \ref{sec:appendix}), we show pairwise correlations between the response quality metrics. \textit{,,Empathy"} is highly correlated with a comprehensive response from the doctor in terms of addressed problems and presence of explanations for each decision. Using the existing user reviews for the annotated doctor responses, in Figure \ref{fig:metric-feedback-corr}, we show how each individual quality measure impacts the reviews. Positive user reviews result in higher overall user satisfaction and retention. It is clear that users are satisfied with empathic doctor responses that include some form of treatment, that comprehensively address their problems and that include explanations. Negative reviews are primarily given to responses with poor grammatical correctness and that misunderstand the platform functionality (e.g., incorrectly using the clarifications functionality or responding with a request for a physical visit to the clinic).

\subsection{Scoring, Recommender and Reconciliation Agents}

To enhance the quality of doctor responses within Dr.Copilot, we developed a multi-agent framework that leverages DSPy \cite{khattab2024dspy}, a modular paradigm for prompt optimization. This framework builds on the response quality measures introduced in the previous section. Dr.Copilot (Figure \ref{fig:evaluation-architecture}) comprises three main components: a \textit{Scoring Agent}, a \textit{Recommender Agent} and, for evaluation, a \textit{Reconciliation Agent}. Instead of using brittle handcrafted prompts or directly fine-tuning the model, we opt for automatic prompt optimization to enable rapid training of specialized agents for accurate scoring across each metric. Additionally, the constraint of a limited labeled dataset poses challenges for fine-tuned models, which may struggle to generalize to new data. In contrast, prompt optimization using DSPy has shown its effectiveness with small labeled datasets, making it a suitable choice.

\noindent \textbf{Scoring Agent.}
This agent evaluates doctor responses based on quality metrics, including empathy, number of problems addressed, grammatical correctness and platform functionality usage. For each metric, we design base prompts to guide the model in assigning scores. In App. \ref{sec:appendix} we provide an example of a base prompt (Listing \ref{lst:prompt_scorer}). To ensure structured and consistent outputs for each metric's scoring model, we use structured LLM outputs \cite{willard2023efficient} present in DSPy, which ensures model responses are in the required score formats. To optimize the scoring agent’s performance, we test three DSPy prompt optimizers: Labeled Few-Shot, Bootstrap Few-Shot, and SIMBA \cite{khattab2024dspy}. For the Labeled Few-Shot optimizer, we select representative examples from the training set, ensuring an equal balance of positive and negative reviews samples. The Bootstrap Few-Shot optimizer extends this by including unlabeled samples, capturing correct model predictions and their reasoning traces, including them in the final scoring prompt. The SIMBA optimizer iteratively refines the base prompt for each metric by checking the model scoring performance against labeled samples and then keeps the best performing prompts. 

\noindent \textbf{Recommender Agent.}
This agent processes the scores from the scoring agent and generates tailored recommendations to improve the doctor’s response. We manually design the prompts for the recommender model to ensure it provides relevant suggestions. The prompts are created to present recommendations not as directives, but as suggestions to improve the response presentation to maximize patient satisfaction. In Listing \ref{lst:prompt_recommender} (App. \ref{sec:appendix}) we provide a prompt example.

\noindent \textbf{Reconciliation Agent.}
To evaluate the effectiveness of the recommendations, we implemented a self-evaluation procedure: a reconciliation agent uses the recommender’s suggestions to revise the initial doctor response, and the revised response is re-scored to measure improvements in quality metrics. This way, we can measure the quality of recommendations and determine whether they lead to an increase in the quality of the doctor’s response. In Listing \ref{lst:prompt_reconciliator} (App. \ref{sec:appendix}) we provide a prompt example.

\subsection{Pretrained Models}
We evaluate three models from the Gemma model family \cite{team2025gemma}: Gemma 12B, Gemma 27B, MedGemma-27B. We select these models for the following reasons. Given the sensitive nature of patient data involved in our study, we cannot use any closed-source models available through an API, such as GPT-4 \cite{openai2024gpt4technicalreport} or Gemini \cite{geminiteam2025geminifamilyhighlycapable}, since these models require data to be transmitted to external servers.

\begin{figure*}[hbt!]
    \centering
    \includesvg[width=1.0\textwidth]{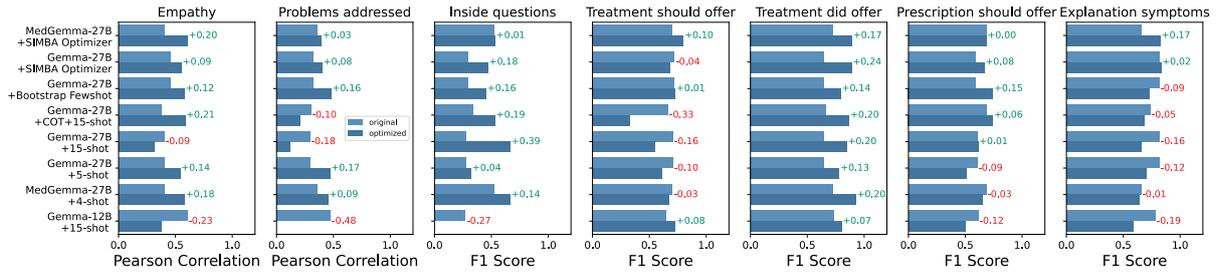}
    \caption{Comparison between base models and prompt-optimized variants across selected quality measures for the \textit{Scoring Agent}. Performance metrics are computed relative to human manual annotations.}
    \label{fig:scoring-comparison}
\end{figure*}

\begin{figure*}[hbt!]
    \centering
    \includesvg[width=1.0\linewidth]{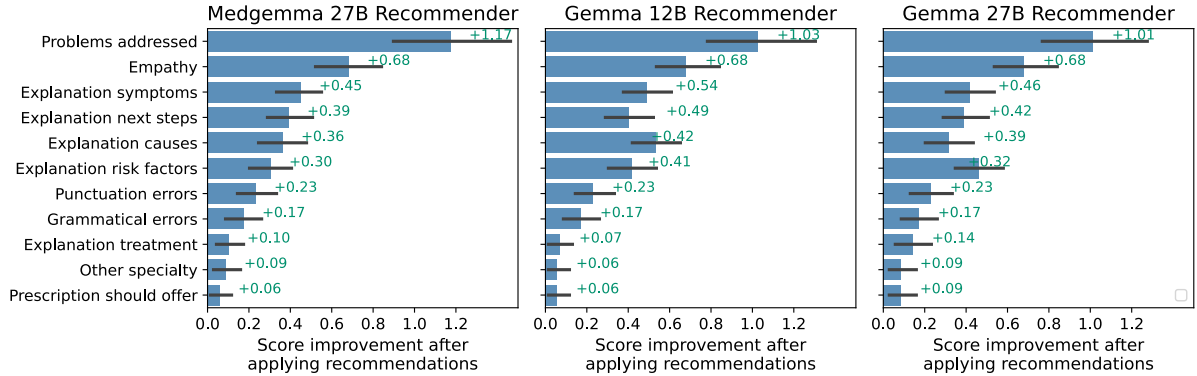}
    \caption{Evaluation of the \textit{Recommender Agent}, by incorporating recommendations through a \textit{Reconciliator Agent} in the original responses and re-scoring.}
    \label{fig:scores-before-after}
\end{figure*}

We turn to open-source alternatives that enable local deployment. Currently, there is no robust model for Romanian with evaluations for the medical domain. Therefore, medium-sized open-weight multilingual models represent a suitable choice. In particular, the Gemma model family has shown robust multilingual capabilities across diverse evaluation benchmarks \cite{team2025gemma}. This ecosystem enables us to evaluate the performance of models with different parameter scales (Gemma 12B, Gemma 27B), alongside models of equivalent size that have different training objectives (Gemma 27B, MedGemma 27B). We detail our experimental configuration in App. \ref{sec:appendix}.

\section{Experiments and Results}
\noindent \textbf{Evaluating the Scorer Agent.}
For evaluating the \textit{Scoring Agent}, in Figure \ref{fig:scoring-comparison} we present a comparison between different base models (Gemma 12B / 27B / MedGemma 27B \cite{team2025gemma}) having the prompt optimized using different optimizers from DSPy \cite{khattab2024dspy}. We computed Pearson Correlations between the output scores and the manual annotations for ,,Empathy" and ,,Problems addressed", and F$_1$ scores for binary measures. For runs having manual few-shots, we selected an equal number of examples from each class, when possible. Overall, MedGemma-27B with SIMBA prompt optimizer has the best overall improvement across all metrics, and we used it for the rest of the experiments.

\noindent \textbf{Evaluating the Recommender Agent.} 
Evaluating free-text recommendations is not trivial, especially since there is no ground truth or human rated examples. A common approach is to use an LLM-as-a-Judge paradigm \cite{gu2024survey} to score responses. However, we propose a \textit{,,Self-Evaluation Procedure"}: we create a \textit{Reconciliation Agent} which is tasked to incorporate the recommendations into the original response, and the revised response is re-scored using the \textit{Scorer Agent}. This setup is closer to real-world deployment, as the \textit{Reconciliation Agent} effectively simulates how doctors use the suggestions to improve the response. In Figure \ref{fig:scores-before-after}, we show the average improvement of metrics after suggestion incorporation and re-scoring by a SIMBA MedGemma-27B scorer. Similarly, we chose MedGemma-27B as a \textit{Recommender Agent}. In the next section, we present results after deployment in a live production environment and show that the improvement estimates of the self-evaluation procedure are actually conservative, and the real score improvement by doctors is greater.

\subsection{Live Deployment Results}

\noindent \textbf{Engagement metrics.}
During the deployment period, Dr.Copilot processed 212 evaluation requests across 41 doctors. A total of 449 recommendations were suggested. In the UI, doctors view recommendations for only the top 3 most important metrics, ranked by their correlation with patient reviews (i.e., Figure \ref{fig:metric-feedback-corr}), to avoid overwhelming them with excessive suggestions. In Fig. \ref{fig:screenshot-deploy} (App. \ref{sec:appendix}) we show the application interface from the doctors' perspective when interacting with Dr.Copilot. In 49 out of the 212 cases, doctors revised their responses based on the recommendations. The distribution of suggested recommendation types is detailed in Table \ref{tab:histogram} (App. \ref{sec:appendix}).

\begin{figure}[hbt!]
    \centering
    \includesvg[width=\linewidth]{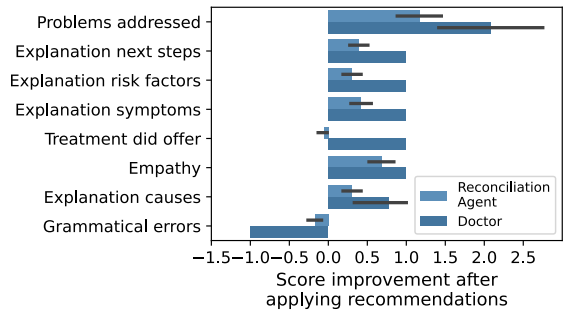}
    \caption{Score improvement across selected metrics by doctors who incorporated suggestions from Dr.Copilot compared to estimations using the \textit{Reconciliation Agent}. Doctors use the feedback better.}
    \label{fig:live-score-improvement}
\end{figure}

\noindent \textbf{Quality improvement.}
In Figure \ref{fig:live-score-improvement}, we show score improvements across metrics when suggestions are incorporated by doctors in the live environment. Compared to the self-evaluation estimation, doctors using Dr.Copilot obtained a 51\% overall relative improvement across metrics, compared to the estimated 37\%.

\noindent \textbf{Patient satisfaction.} 
Patients provide reviews on doctors’ final responses. As the number of questions received varies by doctor due to factors such as availability, reputation and average response quality, we used the like-to-response ratio to assess Dr.Copilot’s impact on perceived patient satisfaction. During the deployment period, responses that incorporated Dr.Copilot’s suggestions received positive reviews from patients in 40.82\% of cases, compared to 23.98\% for responses that did not incorporate the suggestions. This shows a 70.22\% increase in the like-to-response ratio.

\section{Conclusions}
In this work, we introduced Dr.Copilot, a multi-agent LLM system designed to enhance doctor-patient communication quality in Romanian text-based telemedicine settings. Dr.Copilot represents one of the first real-world deployments of Romanian medical LLM applications, overcoming the challenges of working with a low-resource language in a specialized domain. Through automatic prompt optimization using DSPy \cite{khattab2024dspy}, we achieved effective performance with limited labeled data. In a live deployment environment with 41 doctors, we observed a 70.22\% increase in positive patient reviews for responses that incorporated Dr.Copilot's suggestions. The deployment of Dr.Copilot represents an important step forward in the practical application of LLMs in healthcare, particularly for underrepresented languages such as Romanian. 

\section*{Ethical Considerations}
Dr.Copilot is designed with strong ethical safeguards to ensure safe and responsible use in clinical settings. Unlike other autonomous systems that interact directly with patients or provide medical advice \cite{mukherjee2024polaris,liu2025riskagent,zhao2025smart}, Dr.Copilot functions strictly as a supportive tool for physicians. The model does not offer medical recommendations and is not intended to replace professional judgment. Instead, it serves to assist doctors in formulating clearer and more informative responses to patient questions. Furthermore, the system is deployed on-premise, using local and open-weight models \cite{team2025gemma}, without reliance on external API services such as OpenAI or Gemini. This architectural choice minimizes data privacy risks for sensitive health information. The patient questions remain within the institution's infrastructure, and no information is transmitted to third-party services.

\section*{Limitations}

One of the limitations of our work is evaluation with a limited number of doctors. While our results reflect a preliminary evaluation, we expect the results to improve with further iterations. Another limitation of our work is that while the Gemma3 family of models has been evaluated for multilingual capabilities \cite{team2025gemma}, their performance for Romanian and, more specifically, on Romanian medical conversations, is unclear and requires further benchmarking. While we have not observed any particular issues that might arise from language misunderstanding, more comprehensive benchmarks for evaluation on medical domains are needed. 

Any type of (semi-) autonomous system related to the medical domain presents inherent risks. While Dr.Copilot does not offer medical advice, and doctors only interact with it to improve the presentation of their responses, interacting with a chatbot can impact doctors' responses \cite{chatgpt2023isbad} and might result in lost trust from patients if they suspect that responses are algorithmic \cite{reis2024influence}. Furthermore, since Dr.Copilot is a system of multiple LLM agent deployed in a live environment that directly process raw user input (i.e. patient questions), it might be a target of jailbreaking attacks \cite{zeng2024johnny,andriushchenko2024jailbreaking}. This aspect might be mitigated through the use of guardrails \cite{rebedea-etal-2023-nemo} and red-teaming \cite{perez2022red}.

\bibliography{refs}

\appendix
\section{Appendix}
\label{sec:appendix}
\subsection{Doctor Response Evaluation Metrics}

\paragraph{Quality Metrics}
\begin{itemize}
    \item \textbf{Empathy (1-5):} Evaluates emotional consideration in responses
    \begin{itemize}
        \item[1] Harsh response, scolds user, makes negative observations
        \item[2] Abrupt, direct response without emotional consideration, no explanations
        \item[3] Polite but doesn't consider user's emotional state, relatively short with few explanations
        \item[4] Polite response, focuses solely on medical aspects, but does not consider user's emotional state
        \item[5] Empathetic response, considers user's emotional state, explanatory, shows goodwill toward user, attempts to reassure them
    \end{itemize}
    
    \item \textbf{Problems Addressed (1-5):} Completeness of addressing patient concerns
    \begin{itemize}
        \item[1] Doctor addressed none of the patient's problems (e.g., "go see a doctor")
        \item[2] Doctor addressed one main problem, ignoring other questions
        \item[3] Doctor addressed most problems punctually (approximately 80\%)
        \item[4] Doctor addressed all patient problems punctually, without additional completions
        \item[5] Doctor addressed all patient problems, including unknown ones, covering the entire medical act (causes, treatment, prescription, analyses, next steps)
    \end{itemize}
    
    \item \textbf{Grammatical Errors:} Response contains grammatical errors (true/false)
    \item \textbf{Abbreviations:} Response contains abbreviations (true/false)
    \item \textbf{Punctuation Errors:} Response contains punctuation errors (true/false)
\end{itemize}

\paragraph{Treatment Assessment}
\begin{itemize}
    \item \textbf{Treatment Should Offer:} Should the doctor offer treatment in this case? (true/false)
    \item \textbf{Treatment Did Offer:} Did the doctor offer treatment? (true/false)
    \item \textbf{Prescription Should Offer:} For the offered treatment, should the doctor offer a prescription? (true/false)
\end{itemize}

\paragraph{Explanation Completeness}
\begin{itemize}
    \item \textbf{Causes Explanation:} Response mentions causes of the problem (true/false)
    \item \textbf{Symptoms Explanation:} Response mentions symptoms (true/false)
    \item \textbf{Risk Factors Explanation:} Response mentions risk factors (true/false)
    \item \textbf{Next Steps Explanation:} Response explains next steps, including recommendations for where to get analyses (true/false)
\end{itemize}

\paragraph{Response Classification}
\begin{itemize}
    \item \textbf{Questions in Response:} Doctor answered patient's question but addressed additional questions in response (true/false)
    \item \textbf{Other Specialty:} Doctor mentions cannot help patient because case requires another medical specialty (true/false)
    \item \textbf{Only Recommends Visit:} Doctor only recommends physical consultation and offers no other information/explanations (true/false)
    \item \textbf{Cannot Help Online:} Doctor mentions cannot help user with information in online environment
\end{itemize}

\subsection{Experimental details}
We use vLLM \cite{kwon2023efficient} as the model serving framework due to its high throughput for parallel query processing. This is crucial because each scoring and recommendation request will run 17 concurrent model requests (one per quality measure). We use the original, non-quantized models from Huggingface, \textit{gemma-3-12b-it}\footnote{\href{https://huggingface.co/google/gemma-3-12b-it}{hf.co/google/gemma-3-12b-it}, Accessed: 04.07.2025}, \textit{gemma-3-27b-it}\footnote{\href{https://huggingface.co/google/gemma-3-27b-it}{hf.co/google/gemma-3-27b-it}, Accessed: 04.07.2025}, \textit{medgemma-27b-text-it}\footnote{\href{https://huggingface.co/google/medgemma-27b-text-it}{hf.co/google/medgemma-27b-text-it}, Accessed: 04.07.2025}.

A critical factor in the practical deployment of these models is the latency experienced by doctors awaiting recommendations. By leveraging vLLM with asynchronous queries, we achieve an average response evaluation time of five seconds. Minimizing this latency is vital for optimizing user experience, as extended delays may reduce doctors' willingness to adopt the improvement recommendations. 

For prompt optimization, we split the 100 annotated question-response pairs into a 1:5 train-test ratio: 20 training samples, and 80 validation samples as recommended by DSPy for small, high-quality datasets \cite{khattab2024dspy}).

All experiments were run on a cluster node with two NVIDIA A100 GPUs. A full optimization run using the SIMBA optimizer takes two hours, while a labeled or bootstrapped fewshot run takes 15 minutes. We use the default hyper-parameters for each optimizer, aside from few-shot count. For Labeled Few-Shot we used max\_bootstrapped\_demos = 4, max\_labeled\_demos = 16, max\_rounds = 1, max\_errors = 5. For SIMBA we used batch\_size = 16, num\_candidates = 6, max\_steps = 8, max\_demos = 4.

\begin{figure*}[hbt!]
\begin{lstlisting}[label={lst:prompt_scorer}, caption={Scoring prompt example for the Problems Addressed metric, using DSPy signatures.}]
class ProblemAddressingEvaluator(dspy.Signature):
    """Evaluates how well a doctor addressed patient's problems."""
    patient_question = dspy.InputField(desc="Intrebarea pacientului")
    doctor_response = dspy.InputField(desc="Raspunsul doctorului, care va fi evaluat")
    problems_addressed: Literal["1", "2", "3", "4", "5"] = dspy.OutputField(
        desc="""
    Toate Problemele (1-5):
    1 - Doctorul nu a adresat nici una din problemele pacientului, exemple includ raspunsuri precum "mergeti la doctor"
    2 - Doctorul a adresat o problema principala, ignorand celelalte intrebari
    3 - Doctorul a adresat punctual majoritatea (aproximativ 80%) problemelor
    4 - Doctorul a adresat punctual toate problemele pacientului, fara alte completari
    5 - Doctorul a adresat toate problemele pacientului, inclusiv alte necunoscute, acoperind tot actul medical (cauze, tratament, reteta, analize, pasi urmatori)
    """)
\end{lstlisting}
\end{figure*}

\begin{figure*}[hbt!]
\begin{lstlisting}[label={lst:prompt_recommender}, caption={Recommender prompt example for the Problems Addressed metric, using DSPy signatures.}]
class ProblemAddressingRecommender(dspy.Signature):
    """Identifica problemele neadresate din intrebarea pacientului, precizandu-le printr-o lista.
    Recomandarile trebuie sa fie in limba romana si sa nu inceapa cu "Pentru a imbunatati scorul...".
    Recomandarea poate incepe cu "Raspunsul ar putea beneficia de urmatoarele detalii: <lista probleme neadresate, separate pe cate un rand, maxim 4>"
    """
    patient_question = dspy.InputField(desc="Intrebarea pacientului")
    doctor_response = dspy.InputField(desc="Raspunsul doctorului, care va fi evaluat")
    score = dspy.InputField(desc=description_map["problems_addressed"])
    recommendation = dspy.OutputField(desc="Recomandare pentru adresarea completa a problemelor")

\end{lstlisting}
\end{figure*}

\begin{figure*}[hbt!]
\begin{lstlisting}[label={lst:prompt_reconciliator}, caption={Reconciliator prompt example for the Problems Addressed metric, using DSPy signatures.}]
class ReconciliatorSignature(dspy.Signature):
    """Tries to improve the doctor's response using the provided recommendations"""

    patient_question = dspy.InputField(desc="Intrebarea pacientului")
    doctor_response = dspy.InputField(desc="Raspunsul doctorului, care va fi evaluat")
    recommendations = dspy.InputField(
        desc=f"Recomandari pentru a imbunatati raspunsul doctorului"
    )

    modified_response = dspy.OutputField(
        desc=f"Raspunsul doctorului modificat folosind recomandarile"
    )
    
\end{lstlisting}
\end{figure*}

\begin{figure}[hbt!]
    \centering
    \includesvg[width=\linewidth]{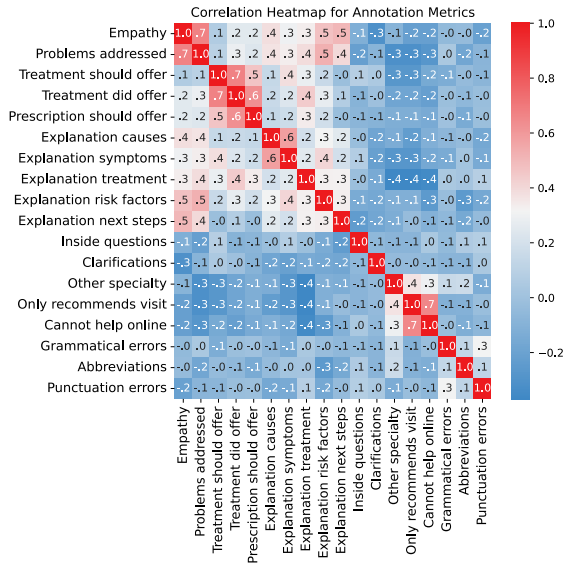}
    \caption{Correlation heatmap between all the quality measures we introduced for doctor responses. For each measure, we used manual annotations fully in agreement after metrics clarification.}
    \label{fig:metric-corr}
\end{figure}

\begin{table}[hbt!]
    \centering
    \resizebox{\linewidth}{!}{
    \begin{tabular}{lrr}
    \textbf{Metric Name} & \textbf{Cohen's Kappa} & \textbf{\% Agreement} \\
    \midrule
    Empathy & 0.69 & -- \\
    Problems addressed & 0.80 & --\\
    Treatment did offer & 0.80 & 90.91 \\
    Treatment should offer & 0.59 & 79.55 \\
    Clarifications & 0.56 & 96.59 \\
    Explanation treatment & 0.55 & 79.55 \\
    Prescription should offer & 0.48 & 80.68 \\
    Other specialty & 0.47 & 93.18 \\
    Explanation causes & 0.41 & 70.45 \\
    Explanation symptoms & 0.37 & 70.45 \\
    Explanation risk factors & 0.34 & 67.05 \\
    Inside questions & 0.30 & 85.23 \\
    Explanation next steps & 0.24 & 61.36 \\
    Cannot help online & 0.24 & 88.64 \\
    Only recommends visit & 0.12 & 88.64 \\
    \end{tabular}
    }
    \caption{Agreement between annotators after the first round of annotation. For \textit{,,Empathy"} and \textit{,,Problems addressed"}, we used a weighted Cohen's Kappa.}
    \label{tab:agreement}
\end{table}

\begin{figure}[hbt!]
    \centering
    \includesvg[width=\linewidth]{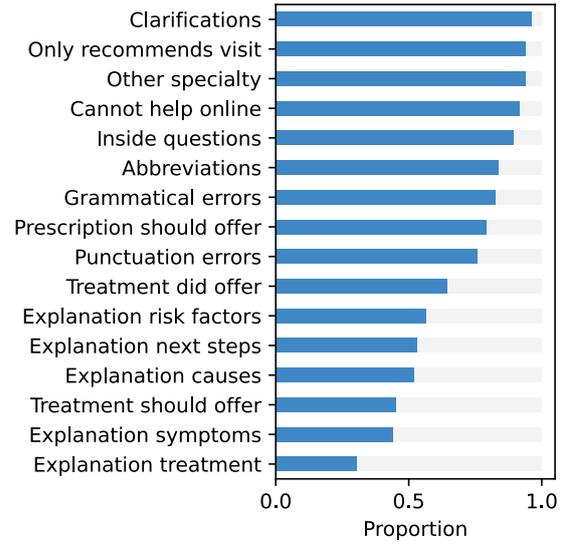}
    \caption{The distribution of across classes for the binary quality measures.}
    \label{fig:metric-proportion}
\end{figure}

\begin{figure}[hbt!]
    \centering
    \includesvg[width=\linewidth]{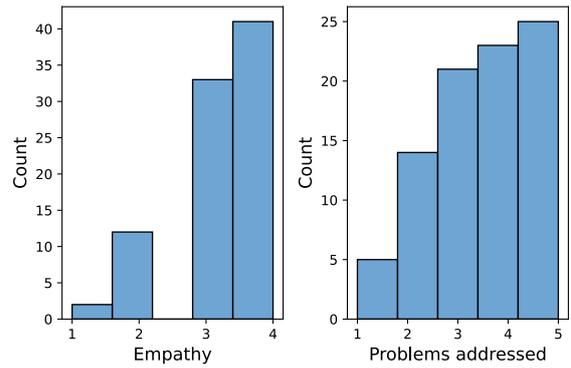}
    \caption{The distribution of values for Likert-style quality metrics.}
    \label{fig:int-metric-proportion}
\end{figure}

\begin{figure}[hbt!]
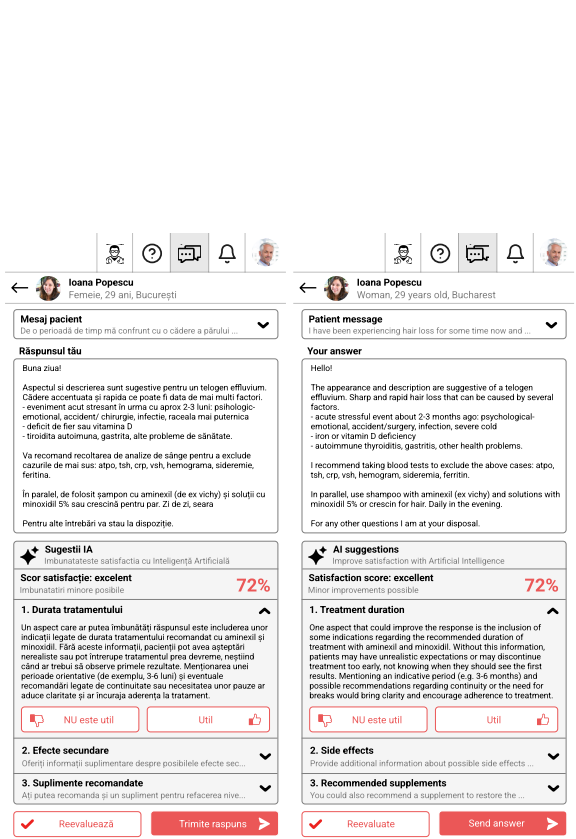

    \centering
    \includesvg[width=0.48\linewidth]{images/screenshot-svg.svg}
    \includesvg[width=0.48\linewidth]{images/screenshot-svg-en.svg}
    \caption{Example of suggestions (in Romanian on the \textbf{left} and the translation in English on the \textbf{right} for illustration purposes) given by Dr.Copilot from the doctors' perspective on the telemedicine platform. The doctor writes the response and then receives targeted feedback for the most important metrics. The names and doctor response are mock-ups, and the profile images are stock photos and do not represent real people. Best viewed on a computer, zoomed in.}
    \label{fig:screenshot-deploy}
\end{figure}

\begin{table}[hbt!]
\centering
\begin{tabular}{l|c}
\textbf{Name} & \textbf{Count} \\
\midrule
Prescription should offer & 95 \\
Explanation causes & 86 \\
Problems addressed & 77 \\
Explanation symptoms & 70 \\
Empathy & 38 \\
Cannot help online & 18 \\
Only recommends visit & 13 \\
Explanation next steps & 13 \\
Explanation risk factors & 12 \\
Other specialty & 8 \\
Treatment did offer & 7 \\
Grammatical errors & 6 \\
Clarifications & 6 \\
\end{tabular}
\caption{Histogram of each recommendation type, during live deployment.}
\label{tab:histogram}
\end{table}

\end{document}